\pdfoutput=1

\documentclass[11pt]{article}

\usepackage[final]{acl}

\usepackage{times}
\usepackage{latexsym}

\usepackage{float}
\usepackage{amsmath}
\usepackage{geometry}
\usepackage{svg}
\usepackage{titlesec}
\usepackage[many]{tcolorbox}
\usepackage{tablefootnote}

\newtcolorbox{boxA}{
    boxrule = 1pt,
    colframe = black, 
    boxrule=0.5pt, width=\textwidth
}

\usepackage[T1]{fontenc}

\usepackage[utf8]{inputenc}
\usepackage{booktabs}

\usepackage{microtype}

\usepackage{inconsolata}

\usepackage{graphicx}

\usepackage{diagbox}

\newcommand*{\affmark}[1][*]{\textsuperscript{#1}}

\title{WisPerMed at ``Discharge Me!'': Advancing Text Generation in Healthcare with Large Language Models, Dynamic Expert Selection, and Priming Techniques on MIMIC-IV}

\author{\bf Hendrik Damm\affmark[1,2],\quad Tabea M. G. Pakull\affmark[3,1],\quad Bahadır Eryılmaz\affmark[4,5],\\ \quad \bf Helmut Becker\affmark[4], \quad \bf Ahmad Idrissi-Yaghir\affmark[1,2],\quad Henning Schäfer\affmark[3,1], \\ \quad \bf Sergej Schultenkämper\affmark[6], and Christoph M. Friedrich\affmark[1,2]\\
\affmark[1] Department of Computer Science, University of Applied Sciences and Arts Dortmund, \\Dortmund, Germany\\
\affmark[2] Institute for Medical Informatics, Biometry and Epidemiology (IMIBE), \\University Hospital Essen, Essen, Germany\\ 
\affmark[3] Institute for Transfusion Medicine, University Hospital Essen, Essen Germany\\
\affmark[4] Institute for AI in Medicine (IKIM), University Hospital Essen, Essen Germany\\
\affmark[5] Institute of Diagnostic and Interventional Radiology and Neuroradiology,\\ University Hospital Essen, Essen, Germany\\
\affmark[6] Bielefeld University of Applied Sciences and Arts, Bielefeld, Germany \\
\texttt {\{hendrik.damm, christoph.friedrich\}@fh-dortmund.de}}

\begin{document}
   \maketitle
\begin{abstract}
This study aims to leverage state of the art language models to automate generating the ``Brief Hospital Course'' and ``Discharge Instructions'' sections of Discharge Summaries from the MIMIC-IV dataset, reducing clinicians' administrative workload. We investigate how automation can improve documentation accuracy, alleviate clinician burnout, and enhance operational efficacy in healthcare facilities. This research was conducted within our participation in the Shared Task Discharge Me! at BioNLP @ ACL 2024. 
Various strategies were employed, including few-shot learning, instruction tuning, and Dynamic Expert Selection (DES), to develop models capable of generating the required text sections. Notably, utilizing an additional clinical domain-specific dataset demonstrated substantial potential to enhance clinical language processing.
The DES method, which optimizes the selection of text outputs from multiple predictions, proved to be especially effective. It achieved the highest overall score of 0.332 in the competition, surpassing single-model outputs. This finding suggests that advanced deep learning methods in combination with DES can effectively automate parts of electronic health record documentation. These advancements could enhance patient care by freeing clinician time for patient interactions. The integration of text selection strategies represents a promising avenue for further research.
\end{abstract}

\section{Introduction}

Clinical notes in electronic health records (EHRs) are used by clinicians to document patient progress in free-text format. These notes typically include the patient's experiences, symptoms, findings, diagnoses, and details of procedures and interventions performed. They serve as the foundation for Discharge Summaries (DS), which contain a section with concise overviews of the entire hospital encounter known as Brief Hospital Course (BHC) \citep{SEARLE2023104358}. They are embedded in the DS and are written by senior physicians who are responsible for the patient's overall care. In addition to BHC, DS also includes Discharge Instructions (DI), which are detailed guidelines provided to patients regarding their post-hospital care. These instructions cover the patient's ongoing care, such as medication instructions, follow-up appointments, and any necessary lifestyle adjustments to ensure proper recovery. Discharge Instructions are designed to facilitate a smooth transition from hospital care to home care and to prevent readmissions. Writing such summaries (BHC) and instructions (DI) can be time-consuming and tedious. Consequently, physicians often spend a big portion of their clinical day dedicated to EHR documentation and desk work \cite{Sinsky2016}.

This paper presents WisPerMed's contribution to the Shared Task Discharge Me! \citep{xu-etal-2024-overview}, which is part of BioNLP @ ACL 2024. This Shared Task aims to ease the administrative burden on clinicians by developing automated methods to generate critical sections in DS, specifically the ``Brief Hospital Course'' and ``Discharge Instruction''. Automating the creation of these sections has the potential to improve documentation accuracy, reduce clinician burnout, and ultimately optimize the processes in healthcare facilities \citep{patel2023chatgpt} by allowing clinicians to allocate more time toward direct patient care.

Our work focuses on designing and implementing various innovative approaches to overcome this challenge and contribute to the overall goals of the Shared Task.

\section{Dataset}

The dataset \citep{xu2024discharge} provided for this Shared Task utilizes the MIMIC-IV (Medical Information Mart for Intensive Care) database \citep{johnson2023mimiciv,Johnson2023}. MIMIC-IV is a publicly available database sourced from the EHR of the Beth Israel Deaconess Medical Center and is accessible on PhysioNet \citep{PhysioNet}. 

The task dataset is divided into four subsets: a training set consisting of 68,785 samples, a validation set containing 14,719 samples, a phase I testing set with 14,702 samples, and a phase II testing set comprising 10,962 samples. Each subset includes DS that are organized into various sections. All records contain two mandatory sections: ``Brief Hospital Course'' and ``Discharge Instructions''. The BHC section typically provides an overview of the patient's treatment and progress during their hospital stay and precedes the DI section. These DI summarize post-hospitalization care instructions and are positioned at the conclusion of the summary.

The challenge organizers provided a regular expression (regex) query to extract these two sections from the DS. The regex query ensures that the relevant information is accurately identified and separated from the rest of the DS content.

For the remainder of this paper, any reference to the ``Discharge Summary'' (DS) will exclude the target sections, BHC or DI.

\section{Evaluation}
The submissions to the Shared Task were evaluated using eight metrics, which assess the relevance and factuality of the generated target. These metrics include Bilingual Evaluation Understudy (BLEU-4) \citep{papineni2002bleu},  Recall-Oriented Understudy for Gisting Evaluation (ROUGE-1, ROUGE-2, ROUGE-L) \citep{lin-2004-rouge}, BERTScore \citep{Zhang*2020BERTScore:}, Metric for Evaluation of Translation with Explicit Ordering (METEOR) \citep{METEOR}, AlignScore \citep{zha-etal-2023-alignscore}, and  Medical Concept (MEDCON) \citep{medcon}. The overall score was calculated by averaging the scores across these eight metrics. In addition to these evaluation metrics, readability scoring metrics were also investigated and utilized in some of the developed approaches. 

After the conclusion of the competition, submissions from the highest-performing teams, determined by the overall score, were evaluated by a panel of clinicians. The generated sections were assessed based on their completeness, correctness, readability, and overall comparison to the reference text. Detailed scoring criteria and evaluation methods used by the clinicians are explained in Appendix \ref{sec:app_clinic}.

\subsection{Relevance}
Relevance was evaluated using  BLEU-4, ROUGE-1, ROUGE-2, ROUGE-L and BERTScore. BLEU-4 measures the precision of 4-gram matches between the generated target and reference text, providing a quantitative measure of how closely the generated target matches the reference in terms of specific sequences of words. The ROUGE metrics measure the overlap of n-grams between the target and reference texts, providing a quantifiable measure of content overlap. Furthermore, BERTScore leverages contextual embeddings to assess the semantic similarity between texts by utilizing pre-trained language models such as BERT \citep{devlin-etal-2019-bert}. In this Shared Task, the distilBERT model \citep{DBLP:journals/corr/abs-1910-01108}, a lightweight and efficient variant of BERT, was used for the BERTScore evaluation.

\subsection{Factuality}
Factuality in text generation was assessed using AlignScore and Summary Consistency (SummaC) \citep{laban-etal-2022-summac}. AlignScore measures how well the facts in a generated summary align with those in the source text. SummaC extends the AlignScore by considering both, the alignment and consistency of the generated target, ensuring it not only contains factual information but also maintains logical coherence with the source.

Furthermore, METEOR score evaluates translation quality by aligning machine-generated target with reference translations, considering synonyms, stemming, and ordering. It balances precision and recall, and penalizes non-contiguous matches to more closely reflect human judgments than simpler metrics like BLEU-4. Lastly, the MEDCON score is a medical concept-based evaluation metric that uses the F1-score to measure the similarity between the Unified Medical Language System (UMLS) concept sets found in candidate and reference clinical notes, assessing their accuracy and consistency. 

\subsection{Readability}
Readability was assessed using the Flesch-Kincaid Grade Level (FKGL) \citep{Kincaid1975DerivationON}, Dale-Chall Readability Score (DCRS) \citep{chall1995}, and Coleman-Liau Index (CLI) \citep{Coleman1975ACR}. FKGL estimates the educational grade level of a text based on sentence length and syllable count per word. DCRS evaluates text complexity by identifying words not recognized by typical fourth graders. CLI calculates the grade level needed to understand the text based on character counts and sentence structure. According to CLI, higher scores indicate lower readability.

\section{Methods}
In this section, different approaches to tackle the Shared Task are described. Licenses for the used models, frameworks, and additional datasets can be found in Appendix \ref{sec:appendix_lic}.

\subsection{Few-Shot learning}
Few-shot learning \citep{fsl} enables machine learning models to quickly adapt to new tasks using only a handful of training examples, reducing the need for extensive data collection. This method has shown improved performance on new tasks with minimal input. The few-shot approach utilized the 
WizardLM-2-8x22B (WizardLM-2) \citep{xu2024WizardLM} model, which was released by Microsoft and is an instruction tuned version of the Mixtral-8x22B\footnote{\url{https://mistral.ai/news/mixtral-8x22b/} \\Accessed: 2024-05-14} model from Mistral AI. Refer to Appendix \ref{sec:appendix} for prompting examples.

\subsection{Instruction Tuning}

The process of instruction tuning \citep{peng2023instruction} in natural language processing involves guiding a pre-trained large language model to follow specific instructions or prompts. Unlike traditional fine-tuning, which focuses on adapting the model to a specific task using a task-specific dataset, instruction tuning uses diverse instruction-based datasets to train the model to generate more accurate and relevant responses to a wide range of queries. This enables the model to better generalize across different tasks by understanding and following the instructions given.

For every experiment carried out, two models were trained: One to generate DI and one to generate BHC. Between the different experiments, hyperparameters were changed only slightly to make the experiments comparable (see Appendix \ref{sec:appendix_param}). As input format, the chat template recommended by the model publishers was used for training. Chat templates\footnote{\url{https://huggingface.co/docs/transformers/chat_templating} Accessed: 2024-05-17} are structured formats that guide the interaction between the user and the model. The input consisted of a System Message and the DS taken from the MIMIC-IV dataset. Example prompts are shown in the Appendix (see Appendix \ref{sec:appendix}). Most models were trained on a single NVIDIA H100 80GB using the unsloth\footnote{\url{https://github.com/unslothai/unsloth} \\Accessed: 2024-05-14} framework. Only Phi-3-Mini-128K-Instruct  \citep{abdin2024phi3} was trained on three NVIDIA H100 80GB. It was necessary to choose Large Language Models that are capable of handling long sequences. The average DS length is about 1,775 words or 4,243 tokens, using the Mistral-7B-Instruct-v0.2 \citep{jiang2023mistral} tokenizer. All models were trained with Low-Rank Adaptation (LoRA) \citep{hu2022LoRA}. 
The following models were evaluated: Llama-3-8B-Instruct \citep{llama3modelcard}, Llama-3-70B-Instruct \citep{llama3modelcard}, OpenBioLLM-70B \citep{openBio}, Phi-3-Mini-128K-Instruct, Mistral-7B-Instruct-v0.2. In the remainder of the paper, ``I'' stands for Instruct in the model naming convention. Please see Appendix \ref{sec:appendix_param} for the fine-tuning setup.

Besides the classical approach of model fine-tuning, an attempt was made to prime the models to improve their understanding of ``clinical language''. For this, the models were instruction-tuned with the Asclepius dataset before using the task-specific MIMIC-IV dataset. For this approach, Llama-8B-I and Mistral-7B-I-v0.2 were evaluated.

Asclepius\label{asclep} is a dataset that was released by \citealt{asclepius}. This dataset contains 158,000 rows of synthetical clinical notes and Instruction-Answer pairs. It was built on publicly available case reports, extracted from biomedical lectures, and then transformed into clinical notes. instruction-answer pairs were built using ChatGPT-3.5-Turbo \cite{chatgpt}.

\subsection{MIMIC Section Identification}
MIMIC Section Identification (MIMIC-SID) \citep{mimicsid,zensols} is a framework used for automatically classifying sections within unstructured clinical texts, such as patient medical records. It recognizes and defines different sections of text based on their content and context. This is particularly useful in the medical domain. Documents such as DS contain distinct sections (e.g., diagnosis, treatment, patient history) that need accurate identification for effective information retrieval and processing.

\begin{figure}[!ht]
  \centering
  \includegraphics[scale=0.7]{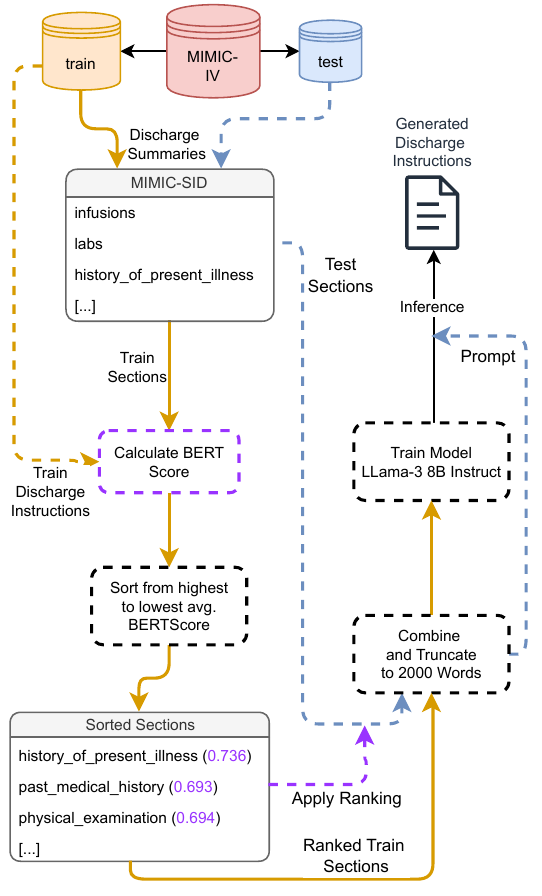}
  \caption{This workflow, exemplified by DI, is applied to BHC in the same way. With MIMIC-SID the dataset is divided into up to 50 sections. For each training section, the average BERTScore is computed using the target text as a reference. The sections are then ranked from highest to lowest BERTScore, and this ranking is applied to both the training and testing DS. The ranked training dataset is used to train the Llama-3-8B-I model. Subsequently, the ranked testing dataset is presented to the model in the form of prompts to generate DI outputs.}
  \label{fig:ft_wf}
\end{figure}

Utilizing MIMIC-SID (see Figure \ref{fig:ft_wf}), the most important sections for the target text were identified by calculating the average BERTScore (with distilBERT) between the extracted section and the target section. The text was then ordered based on relevance, from highest to lowest BERTScore, and truncated after 2,000 words. This method assumes that relevant parts are already found at the beginning of the text, and less relevant parts would be cut out. To compare this approach to a more standardized setting, the unaltered input text was also truncated to 2,000 words. This results in two training schemes: one with 2,000 words of reordered text and one with 2,000 words of the original text.

\subsection{Hyperparameters}
\label{sec:inf}

The quality of the generated targets is strongly influenced by the inference parameters employed. The Meta-Llama-3-8B-Instruct model was utilized to establish decoding strategies for the Shared Task, specifically adopting the proposed methods by \citep{minaee2024large}. Three experimental runs were conducted to examine their influence on text generation quality, each employing these decoding strategies in different configurations. The configurations and their respective parameters are detailed in Table~\ref{tab:llm-configurations}.

\begin{table}[h!]
\centering
\begin{tabular}{llll}
\toprule
\textbf{Parameter} & \textbf{Config 1} & \textbf{Config 2} & \textbf{Config 3} \\ 
\midrule
do\_sample & False & False & True \\ 
RP & 1.2 & 1 & 1 \\ 
NNS & 3 & $\infty$ & $\infty$ \\ 
ERP & 1 & 1.2 & 1 \\ 
temp & 0 & 0 & 0.6 \\ 
top\_p & 0 & 0 & 0.9 \\ 
\bottomrule
\end{tabular}
\caption{Configuration Parameters for Inference Runs. RP stands for \textit{repetition\_penalty}, NNS stands for \textit{non\_repeat\_ngram\_size}, and ERP stands for \textit{endocer\_repetition\_penalty}.}
\label{tab:llm-configurations}
\end{table}

\subsection{Dynamic Expert Selection}
As final approaches, five different Dynamic Expert Selections (DES) were constructed. For each DES, a set of models was pre-selected to serve as experts. Each model generates DI and BHC for a DS, and then an expert model is selected whose text is included in the submission.

Readability and factuality scores are calculated to select the expert model. The readability scores can be calculated without any reference text, and the factuality scores use the entire DS as a reference. Because they do not use the target texts, these scores are referred to as pre-calculated scores.

\begin{figure}[t]
\centering
  \includegraphics[scale=0.4]{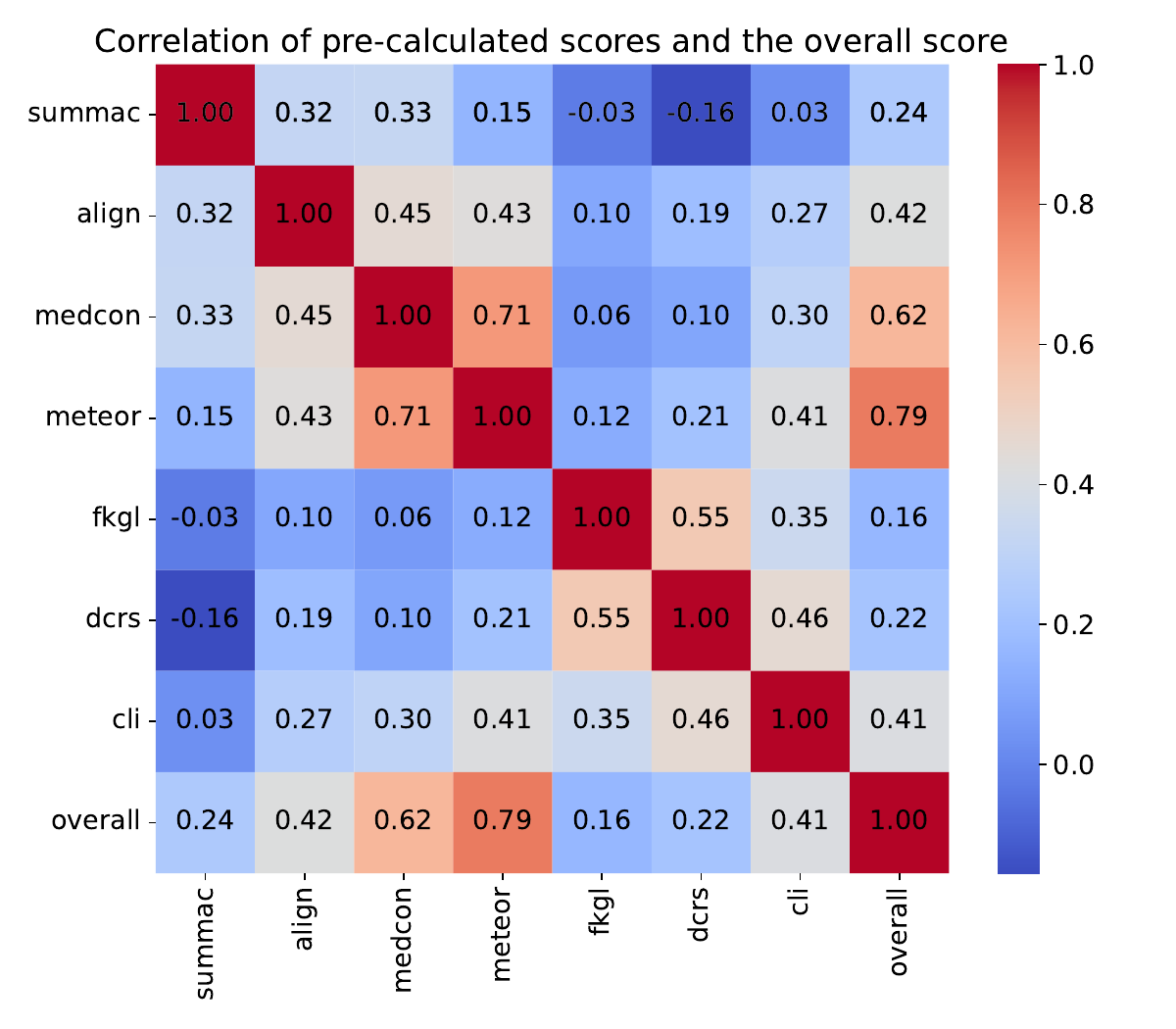}
  \caption{Heatmap of the Pearson correlations between pre-calculated scores and the overall score on the validation dataset. The pre-calculated scores include factuality scores (SummaC, AlignScore, MEDCON and METEOR), which are calculated for the generated targets of the Mistralv2 + Asclepius model with the whole DS as the reference, and readability scores (FKGL, DCRS and CLI).}
  \label{fig:corr}
\end{figure}

Additionally, the validation set was used to compute the pre-calculated scores for the generated targets of the Mistral-7B-I-v0.2 + Asclepius model. Furthermore, the overall scores (all challenge evaluation scores) based on the target texts were determined on the validation set. Then, the correlations between the pre-calculated scores and the overall scores were examined. Figure \ref{fig:corr} shows these correlations as a heatmap. Taking into account these correlations, DES 1-4 were constructed. For DES 5, the lengths of the generated targets were considered instead of scores as the selection criterion. This decision was based on the observation that, particularly in longer texts, models exhibit signs of hallucination or the generation of repetitive content.

When compiling a DES, the pre-calculated scores of all included models for a DS are subjected to a min-max normalization. This means normalization over all available models. These normalized scores are then multiplied by selected weights. The model with the highest average of all normalized and weighted scores is selected as the expert for that DS. 

\begin{figure}
    \begin{tcolorbox}[boxrule=0.5pt, colframe=black,coltitle=white, colbacktitle=darkgray, width=\columnwidth, title=Example given:]
       {\footnotesize 
[...Non-repetitive text]\\
\textit{
-Please check LFTs weekly for the next month. \\
-Please continue to monitor for signs of refeeding 
hyperglycemiAsclepius\\
-Please continue to monitor for signs of refeeding 
hypocalcemiAsclepius\\ }
[Continues Repetition...]}
    \end{tcolorbox}
    \caption{Example of repetitive and hallucinated DI output generated by Llama-3-8B-I. The words hyperglycemia and hypocalcemia are very similar but only one of them should be in the generated targets. The other one was not mentioned in the DS.}
\end{figure}

\paragraph{DES 1}
This DES was optimized for MEDCON and METEOR with a weight of \(\frac{1}{2}\) each, as these two metrics exhibited the strongest correlation with a higher overall score. The final submission file included 6,407 texts from Mistral-7B-I-v0.2 + Asclepius, 1,210 texts from Llama-3-8B-I with greedy decoding \citep{minaee2024large}, 2,815 texts from Llama-3-8B + Asclepius, 5,600 texts from Llama-3-70B-I, 4,112 from OpenBioLLM-70B, and 1,780 from WizardLM-2.
\paragraph{DES 2}
This DES was optimized for MEDCON and METEOR as measures for factuality, and CLI score for readability with the weights \(\frac{2}{5}\) for both MEDCON and METEOR, and \(\frac{1}{5}\) for CLI. The final submission file included a total of 3,471 texts from the Mistral-7B-I-v0.2 + Asclepius model, 4,374 texts from Llama-3-8B-I + Asclepius, 5,108 texts from Llama-3-70B-I, and 5,971 texts from OpenBioLLM-70B.
\paragraph{DES 3}
This DES was optimized for all readability metrics (FKGL, DCRS, and CLI) for DI, and additionally MEDCON, METEOR, and AlignScore as factuality metrics for both text types. For the DI, all readability metrics were assigned a weight of \(-\frac{1}{9}\), and the factuality metrics were assigned a weight of \(\frac{2}{9}\). For the BHC, all factuality metrics were weighted with \(\frac{1}{3}\). The final submission file included 5,120 texts from Mistral-7B-I-v0.2 + Asclepius, 4,211 texts from Llama-3-8B-I + Asclepius, 5,435 texts from Llama-3-70B-I, and 7,158 texts from OpenBioLLM-70B.
\paragraph{DES 4}
This DES used only Mistral-7B-I-v0.2 models and the values of correlation between the scores calculated on the generated targets against the whole DS and the overall score on the validation dataset as weights. The final submission file included 7,912 texts from Mistral-7B-I-v0.2 + Asclepius, 6,485 texts from the Mistral-7B-I-v0.2 + Asclepius model, which was further fine-tuned on the validation dataset, and 7,527 texts from Mistral-7B-I-v0.2.
\paragraph{DES 5}
This DES considers lengths of texts instead of weighted metrics. DI in the training dataset has an average word count of approximately 196.3, whereas BHCs have an average word count of approximately 327.6. Models trained on these texts, therefore, tend to generate shorter texts for DI and longer texts for BHC. To mitigate the impact of hallucinations at the end of lengthy texts, a strategy of preferably selecting shorter texts for the DES was adopted. The objective of the strategy was to initially rank the models based on their overall scores. Subsequently, for each DS, the text from the first model that has a word count within the range of 100 to 180 words is selected. If no model had generated a text with a word count within this range, the text with the minimum word count was selected. However, the text could not be shorter than 70 words. In the case that no text met these criteria, the text from the highest-ranked model remained.

\section{Results}
The final scores for the Shared Task, provided by the organizers, are presented in Table \ref{tab:results}. The table organizes the approaches from the most general to the most specific, beginning with the baseline model, followed by the Few-Shot model, the instruction-tuned models, instruction-tuned models primed with Asclepius, MIMIC section-based approaches, and various DES variants. The top competitors by overall score are highlighted for comparison purposes. All inference runs for the final results were conducted with an optimized decoding (temp=0.6, top\_p=0.9) strategy as described in Section \ref{sec:inf}. Based on its high evaluation scores, Configuration 3 was chosen as the standard parameter setting, which can be seen in Table \ref{tab:llm-configurations}. \\
\begin{table*}[ht!]
\centering
\begin{tabular}{ l  l  l l  l  l  l  l  l  l}  
\toprule
\textbf{Model}& \textbf{Ovr.} &  \textbf{BLEU} & \textbf{R-1}& \textbf{R-2}& \textbf{R-L}& \textbf{BERT}& \textbf{MET} & \textbf{Align} & \textbf{MED}\\
\midrule
\multicolumn{10}{c}{Baseline}\\
\midrule
Challenge Baseline& 0.102 & 0.015 & 0.126 & 0.051 & 0.113 & 0.138 & 0.098 & 0.167 & 0.121\\
\midrule
\multicolumn{10}{c}{Few-Shot learning}\\
\midrule
WizardLM-2 8x22B& 0.195 & 0.017 & 0.257 & 0.074 & 0.158 & 0.331 & 0.310 & 0.193 & 0.218\\
\midrule
\multicolumn{10}{c}{instruction tuned}\\
\midrule
Llama-3-8B-I & 0.253 & 0.053 & 0.331 & 0.107 & 0.241 & 0.392 & 0.235 & 0.320 & 0.348 \\
Mistral-7B-I-v0.2  & 0.289 & 0.101 & 0.371 & 0.122 & 0.252 & 0.416 & \textbf{0.375} & 0.293 & 0.380 \\
Llama-3-70B-I & \textbf{0.300} & \textbf{0.112} & 0.367 & \textbf{0.141} & \textbf{0.260} & \textbf{0.437} & 0.347 & 0.334 & \textbf{0.401} \\
OpenBioLLM-70B & 0.285 & 0.084 & 0.376 & 0.127 & 0.248 & 0.421 & 0.307 & \textbf{0.337} & 0.383 \\
Phi-3-mini-128k-I & 0.254 & 0.062 & \textbf{0.347} & 0.128 & 0.217 & 0.359 & 0.310 & 0.275 & 0.330 \\
\midrule
\multicolumn{10}{c}{instruction tuned + Asclepius (A.)}\\
\midrule
Llama-3-8B-I + A. & 0.302 & 0.107 & 0.388 & \textbf{0.150} & \textbf{0.275} & 0.432 & 0.350 & 0.311 & 0.403\\
Mistral-7B-I-v0.2 + A. & \textbf{0.307} & \textbf{0.120} & \textbf{0.390} & 0.140 & 0.258 & \textbf{0.434} & \textbf{0.391} & \textbf{0.320} & \textbf{0.404}\\
\midrule
\multicolumn{10}{c}{MIMIC Section Identification}\\
\midrule
Llama-3-8B-I 2k & 0.209 & 0.022 & 0.263 & 0.054 & 0.171 & 0.326 & \textbf{0.199} & \textbf{0.355} & 0.280\\
Llama-3-8B-I R 2k  & \textbf{0.216} & \textbf{0.026} & \textbf{0.292} & \textbf{0.073} & \textbf{0.191} & \textbf{0.351} & 0.186 & 0.306 & \textbf{0.304}\\
\midrule
\multicolumn{10}{c}{Hyperparameter}\\
\midrule
Llama-3-8B-I Greedy & 0.192 & 0.018 & 0.274 & 0.043 & 0.147 & 0.314 & \textbf{0.221} & 0.281 & 0.241\\
Llama-3-8B-I ERP & \textbf{0.238} & \textbf{0.032} & \textbf{0.348} & \textbf{0.093} & \textbf{0.228} & \textbf{0.372} & \textbf{0.221} & \textbf{0.307} & \textbf{0.300}\\
\midrule
\multicolumn{10}{c}{Dynamic Expert Selection}\\
\midrule

DES 1 & 0.277 & 0.097 & 0.329 & 0.121 & 0.217 & 0.417 & 0.339 & 0.319 & 0.374\\

DES 2 & 0.311 & 0.110 & 0.414 & 0.151 & 0.273 & \underline{\textbf{0.439}} & 0.351 & \underline{\textbf{0.344}} & 0.406\\

DES 3 & 0.296 & 0.108 & 0.366 & 0.128 & 0.242 & 0.435 & 0.352 & 0.335 & 0.400\\

DES 4 & 0.297 & 0.112 & 0.371 & 0.127 & 0.244 & 0.426 & 0.379 & 0.320 & 0.396\\

DES 5 & \underline{\textbf{0.332}} & \underline{\textbf{0.124}} & \underline{\textbf{0.453}} & \underline{\textbf{0.201}} & \underline{\textbf{0.308}} & 0.438 & \underline{\textbf{0.403}} & 0.315 & \underline{\textbf{0.411}}\\
\midrule
\multicolumn{10}{c}{Top 5 Competitors}\\
\midrule
HarmonAI Lab Yale & \textbf{0.300} & \textbf{0.106} & 0.423 & 0.180 & \textbf{0.284} & \textbf{0.412} & 0.381 & 0.265 & 0.353 \\
aehrc & 0.297 & 0.097 & 0.414 & \textbf{0.192} & \textbf{0.284} & 0.383 & \textbf{0.398} & 0.274 & 0.332 \\
EPFL-MAKE & 0.289 & 0.098 & \textbf{0.444} & 0.155 & 0.262 & 0.399 & 0.336 & 0.255 & \textbf{0.360} \\
UF-HOBI & 0.286 & 0.102 & 0.401 & 0.174 & 0.275 & 0.395 & 0.289 & \textbf{0.296} & 0.355 \\
de ehren & 0.284 & 0.097 & 0.404 & 0.166 & 0.265 & 0.389 & 0.376 & 0.231 & 0.339 \\
\bottomrule
\end{tabular}
\caption{\label{valid-results}Summary of model performance across different experimental settings. Each section represents a distinct approach: Baseline, Few-Shot Learning, instruction tuned, instruction tuned + Asclepius, MIMIC-SID (2k for truncation to 2k words and R for reordering the subsections in the text from most to least relevant according to BERTScore), Hyperparameter, and DES, showcasing respective strategies to address the challenge. \textit{I} indicates that the instruction version of the model was used. Furthermore, the Top 5 runs from other challenge participants are included. Metrics include overall score (Ovr.), BLEU-4 (BLEU), ROUGE-1 (R-1), ROUGE-2 (R-2), and ROUGE-L (R-L), BERTScore (BERT), METEOR (MET), AlignScore (Align) and MEDCON (MED). Bold scores indicate the best performance in each category, with underlined bold scores highlighting the top overall scores across all experiments.}
\label{tab:results}
\end{table*}
WizardLM-2, despite not being fine-tuned on the training data, surpassed the baseline with an overall score of 0.195.

Among the fine-tuned models, Llama-3-70B-I led with a score of 0.300, followed by Mistral-7B-I-v0.2 at 0.289, which has a smaller architecture yet outperformed several larger models, including Llama3-8B-I. Even though the OpenBioLLM-70B has been adapted for clinical use, it underperformed when compared to other models. The Phi-3-128-I model matched the performance of larger models, such as the Llama-8B-I, demonstrating the efficiency of smaller architectures.

Among the configurations, incorporating the Asclepius dataset (see Paragraph \ref{asclep}) into Mistral-7B-I-v0.2 made it clearly outperform Llama3-8B-I. Excluding the DES approaches, this combination achieved the highest performance of all models.

The MIMIC-SID approaches with a shorter context length of 2,000 words displayed weak performances. 

Different parameters on the Llama-8B-I, including greedy decoding and an ERP, yielded lower scores compared to setups utilizing sampling and temperature adjustments.

In the Dynamic Expert Selection category, DES 1 focused on MEDCON and METEOR scores but did not surpass the individual fine-tuned models. DES 2 achieved the highest BERTScore and AlignScore, which represent relevance and factuality, respectively. Reaching an overall score of 0.311, this DES outperformed all individual fine-tuned models. DES 3, aimed at lowering readability metrics, scored 0.296, performing better than DES 1 but lagging behind others. DES 4, using correlation values for optimization, showed negligible improvements. DES 5 achieved the highest overall score of 0.332 and topped the leaderboard by limiting text length. The approach achieved the highest scores in all metrics, except for a slightly lower BERTScore and AlignScore.

\section{Discussion}
Despite being one of the less robust models evaluated, WizardLM-2 exceeded the established baseline, showing its effectiveness in a few-shot learning context. With minimal training examples, the model still produced high-quality texts, according to the metrics, highlighting the potential of few-shot learning in enhancing performance metrics.

The performance of instruction-tuned models revealed mixed outcomes. Despite being a specialized adaptation of the Llama-3-70B-I model tailored for medical contexts, OpenBioLLM-70B underperformed in relation to its base model. This behavior was unexpected, considering its design to enhance relevance and accuracy in clinical applications. Conversely, the Mistral-7B-I-v0.2 model demonstrated impressive capabilities, outperforming both the larger OpenBioLLM-70B and the Llama-8B-I models. This highlights the effectiveness of Mistral-7B-I-v0.2 in handling complex medical text generation and summarization tasks despite its smaller size. In contrast to its reputation as one of the most promising state-of-the-art open-source LLMs, the Llama-3 models have been found to be less effective in this challenge. This is on par with the findings from LMSYS chatbot arena \citep{chiang2024chatbot} where LLama-3 models showed the weakest performance compared to other state-of-the-art models on the task of summarization \citep{llama3arena2024}.

Using the Asclepius dataset for priming substantially improved model performance during the fine-tuning phases. For instance, the Llama-8B-I model's score rose from 0.253 to 0.302, and the Mistral model's performance increased from 0.289 to 0.307. Notably, the Mistral-7B-I-v0.2 + Asclepius model was the top performer in the challenge, aside from DES approaches. This underscores the benefits of further training models with specialized datasets to enhance accuracy and relevance in domain-specific tasks.

The reordering of sections within the MIMIC-SID approach moderately enhanced overall model performance, demonstrating that prioritizing the most relevant sections can be beneficial. However, it is important to note that metrics sensitive to text order, such as METEOR and AlignScore, experienced a decline. This suggests that while reordering can improve general outcomes by emphasizing key information, it may simultaneously compromise the sequential integrity of the text. Therefore, this strategy confirms the utility of structurally optimizing input for task-specific relevance, albeit with some trade-offs in textual coherence.

Exploration of hyperparameter settings revealed that more complex configurations did not yield superior results. The basic approach, utilizing \textit{do\_sample=True}, \textit{temp=0.6}, and \textit{top\_p=0.9}, consistently outperformed other tested configurations, including those with greedy decoding and encoder repetition penalties. This emphasizes the efficacy of maintaining simpler hyperparameter settings for stable and high-quality text generation. Additional complexity in parameter tuning did not always correlate with improved model performance.

The DES that relied on the pre-calculated scores had varying effects on the metrics evaluated. Using MEDCON and METEOR in combination with CLI improved the results, whereas choosing the correlation as weights resulted in no improvement. A possible reason might be that the pre-calculated scores were only calculated on the entire DS and not on the target text, as in the final evaluation. It may also be that the correlations are not always sufficient, and a more elaborate association analysis is needed.

Consequently, the best overall score of all DES was achieved by the approach limiting the text length, suggesting that hallucinations and repetitive sequences have a measurable impact on text quality.

\section{Conclusion}
The research identified several opportunities for future investigation that may enhance the performance and utility of the discussed models. Initially, due to the extensive size of the training dataset and the constraints imposed by the context length of input texts, each model was trained for a maximum of only three epochs. Therefore, extending the training duration may provide improvements and merits further exploration.

Moreover, alterations to inference parameters have demonstrated notable effects on model outputs. For example, employing the ERP parameter, while maintaining other settings constant resulted in a degradation of performance metrics (from 0.253 to 0.238 overall score). This suggests that a systematic evaluation of inference parameters could further enhance model output.

Additionally, the practice of priming the model has substantially improved results. Investigating additional datasets for priming purposes could further optimize model performance and expand its applicability across diverse textual tasks. This could be a promising direction for future research efforts. Further opportunities lie in optimizing section reordering to balance task-specific relevance while maintaining text coherence.

The winning approach in the competition, a DES, achieved the highest overall score, suggesting that generating multiple outputs and developing methodologies to select the optimal text may further improve performance. Therefore, exploring various DES techniques and selection criteria is a field for further research.

Lastly, efforts to enhance the quality of medical machine learning algorithms are ongoing, along with a responsibility to report the environmental impact of the research. In this study, the total energy consumption for training and inference is estimated with 1,552.10 kWh, resulting in 591.35 kg CO\textsubscript{2} emission. Detailed information is provided in Appendix \ref{sec:appendix_env}.

\onecolumn
\appendix
\section{Few-Shot Learning Prompts}
\label{sec:appendix}

This section showcases how the WizardLM-2-Model was instructed. A variety of prompts were tested, and the displayed ones (see Figure \ref{fig:fsl_di} and Figure \ref{fig:fsl_bhc}) yielded the best results, as measured by human evaluation. Detailed instructions were provided for the DI text generation, whereas the details for the BHC generation were excluded. The BHC texts are considerably longer on average and do not follow the same pattern most of the time.

\begin{figure}[H]
    \begin{tcolorbox}[boxrule=0.5pt, colframe=black,coltitle=white, colbacktitle=darkgray, width=\textwidth, title=Discharge Instructions Prompt]
    \footnotesize 
    
    USER: Generate a detailed discharge instruction based on the provided summary, adhering to the style of the provided examples. The instruction should comprehensively cover all aspects of the patient's care, with a total length of about 300-500 words.\\
    
    Please follow the format used in previous discharge instructions:
    \begin{enumerate}
        \item Start with a polite greeting and an expression of gratitude or pleasure for having taken care of the patient.
        \item Describe the reason for hospitalization succinctly.
        \item Detail what occurred during the stay, including any treatments administered, patient responses, and significant changes to the patient's condition.
        \item Outline clear follow-up care instructions, including medications, dietary recommendations, activity level, and scheduled follow-up visits.
        \item Close with a kind farewell and additional well-wishes or reminders.
    \end{enumerate}
    Discharge Instruction Format Example:\\
    Dear [Patient Name],\\
    It was a pleasure taking care of you during your hospitalization at [Hospital Name].\\
    
    Why were you hospitalized?\\
    - [Brief reason for hospitalization]\\
    
    What happened while you were in the hospital?\\
    - [Key details about treatment and patient response]\\
    - [Any significant tests and their results]\\
    - [Any changes to patient condition]\\
    
    What should you do after you leave the hospital?\\
    - [Medications and dosage]\\
    - [Dietary instructions]\\
    - [Activity recommendations]\\
    - [Follow-up appointments]\\
    
    We wish you the best in your recovery!\\
    
    Sincerely,\\
    Your [Hospital Team Name] Team\\
    
    Discharge Instruction Example 1 start:
    
    [Discharge Instruction Example from Validation set]
    
    Discharge Instruction Example 1 end.
    
    [$\cdots$]
    
    Discharge Instruction Example 10 start:\newline
    [Discharge Instruction Example from Validation set]\newline
    Discharge Instruction Example 10 end.\newline
    Discharge Summary:\newline
    [Discharge Summary without target section]\newline
    Start with the Discharge Instructions for the Discharge Summary.\newline
    ASSISTANT:
    \end{tcolorbox}
    \caption{Discharge Instruction Prompt for Few-Shot learning with WizradLM-2.}
    \label{fig:fsl_di}

\end{figure}

\begin{figure}[h]
     \begin{tcolorbox}[boxrule=0.5pt, colframe=black,coltitle=white, colbacktitle=darkgray, width=\textwidth, title=Brief Hospital Course Prompt]
    {
    \footnotesize        
        USER: Here are some Example Brief Hospital Courses.
        
        \vspace{1em}
        
        Brief Hospital Course Example 1 start:
        
        [Brief Hospital Course Example from Validation set]
        
        Brief Hospital Course Example 1 end.

        [$\cdots$]

        Brief Hospital Course Example 7 start:
        
        [Brief Hospital Course Example from Validation set]
        
        Brief Hospital Course Example 7 end.

        \vspace{1em}
        
        Now create a Brief Hospital Course in the same style as in the Examples with the information from the following Discharge Summary:
        
        [Discharge Summary without target section]
        
        ASSISTANT:}
        
    \end{tcolorbox}
    \caption{Brief Hospital Course Prompt for Few-Shot learning with WizardLM-2.}
        \label{fig:fsl_bhc}

\end{figure}

\section{Instruction Tuning Prompts}
\label{sec:appendix_ft}
Figure \ref{fig:instruction tuning prompts} and Figure \ref{fig:instruction tuning prompts2} show the prompts used for instruction tuning DI and BHC. The only difference between the instruction tuning and inference prompts is that the [Target Discharge Instructions] or [Target Brief Hospital Course] was left empty for inference. For each model, the recommended chat template provided by the model inventors was followed and applied. This is especially important when using the instruction version of those models.

\begin{figure}[H]
    \begin{tcolorbox}[boxrule=0.5pt, colframe=black,coltitle=white, colbacktitle=darkgray, width=\textwidth, title=Discharge Instructions Prompt]
      {\footnotesize

        <SYSTEM>You are in the world's best hospital as the best doctor. You're given a patient's details summarized by your medical staff in 'Summary'. You now need to figure out the 'Discharge Instructions' for the patient. Think carefully without error, since you might endanger a patient's life, which we do not want to happen.
        
        \vspace{1em}
        
        <User>Summary: [Discharge Summary without target section]
        
        \vspace{1em}
        
        Discharge Instructions:\\
        <ASSISTANT>[Target Discharge Instructions]
}
\end{tcolorbox}
        
    \caption{Instruction Tuning and Inference Prompt for Discharge Instructions.}
    
    \label{fig:instruction tuning prompts}
\end{figure}

\begin{figure}[H]
    \begin{tcolorbox}[boxrule=0.5pt, colframe=black,coltitle=white, colbacktitle=darkgray, width=\textwidth, title=Brief Hospital Course Prompt]
      {\footnotesize
        <SYSTEM>You are in the world's best hospital as the best doctor. You're given a patient's details summarized by your medical staff in 'Summary'. You now need to figure out a 'Brief Hospital Course' for the patient. Think carefully without error, since you might endanger a patient's life, which we do not want to happen.

        \vspace{1em}
        <USER>Summary: [Discharge Summary without target section]

        \vspace{1em}
        Brief Hospital Course:\\
        <ASSISTANT>[Target Brief Hospital Course]

    }
    \end{tcolorbox}
        
    \caption{Instruction Tuning and Inference Prompt for Brief Hospital Course.}
    
    \label{fig:instruction tuning prompts2}
\end{figure}

\normalsize

\pagebreak
\section{Parameter Setup}
\label{sec:appendix_param}
Whenever possible, hyperparameters were only changed slightly to ensure high comparability between results. The LoRA setup is detailed in Table \ref{tab:LoRA}. The following modules were targeted with LoRA: ``q\_proj'', ``k\_proj'', ``v\_proj'', ``o\_proj'', ``gate\_proj'', ``up\_proj'', and ``down\_proj''. While it is suggested\footnote{\url{https://magazine.sebastianraschka.com/p/practical-tips-for-finetuning-llms} Accessed: 2024-05-17} to use a LoRA Rank = LoRA Alpha * 2, this approach was not chosen due to VRAM efficiency considerations.
\begin{table}[h]
\centering
\begin{tabular}{llllllll}
\toprule
Model & LR &LA & loadIn4Bit & LD & GC & DT \\ 
\midrule
Llama-3-8B-I + A. Prime  & 16 & 16 & true & 0 & true & bfloat16 \\
Mistral-7B-I-v0.2 + A. Prime  & 16 & 16 & true & 0 & true & bfloat16\\
Llama-3-8B-I   & 16 & 16 & true & 0 & true & bfloat16\\
Mistral-7B-I-v0.2   & 16 & 16 & true & 0 & true & bfloat16\\
Llama-3-70B-I   & 16 & 16 & true & 0 & true & bfloat16\\
OpenBioLLM-70B   & 16 & 16 & true & 0 & true & bfloat16\\
Phi-3-mini-128k-I   & 16 & 16 & true & 0 & true & bfloat16\\
Llama-3-8B-I 2k + A   & 16 & 16 & true & 0 & true & bfloat16\\
Mistral-7B-I-v0.2 + A.   & 16 & 16 & true & 0 & true & bfloat16\\
Llama-3-8B-I 2k   & 16 & 16 & true & 0 & true & bfloat16\\ 
Llama-3-8B-I R 2k    & 16 & 16 & true & 0 & true & bfloat16\\
\bottomrule
\end{tabular}
\caption{LoRA Setup for fine-tuning. LR means LoRARank, LA means LoRAAlpha, LD means LoRADropout, GC means Gradient Checkpointing, DT means dtype. By A. Asclepius is meant. Prime means the instruction tuning runs with Asclepius.}
\label{tab:LoRA}
\end{table}
For the detailed training setup, please see Table \ref{tab:trainer}. All models were trained on 80GB H100s and 48GB RTX6000s. The Unsloth open-source training framework was used because it reduced VRAM usage by at least 50\% and subsequently made fine-tuning runs twice as fast. This efficiency allowed training almost all models on a single GPU.
\begin{table}[h]
\centering
\begin{tabular}{llllllllll}
\toprule
Model & MSL & E & GAS & WS & LR & BS & O & S & WD\\ 
\midrule
Llama-3-8B-I + A. P.  & 15,000 & 1 & 4 & 5 & 2e-4 & 4 & adamw\_8bit  & linear & 0.01\\
Mistral-7B-I-v0.2 + A. P.   & 15,000 & 1 & 4 & 5 & 2e-4 & 4 & adamw\_8bit & linear & 0.01 \\

Llama-3-8B-I   & 10,000 & 3 & 4 & 5 & 2e-4 & 4 & adamw\_8bit & linear & 0.01 \\
Mistral-7B-I-v0.2   & 10,000 & 3 & 4 & 5 & 2e-4 & 4 & adamw\_8bit & linear & 0.01 \\

Llama-3-70B-I   & 10,000 & 2 & 4 & 5 & 2e-4 & 2 & adamw\_8bit & linear & 0.01 \\
OpenBioLLM-70B   & 10,000 & 2 & 4 & 5 & 2e-4 & 2 & adamw\_8bit & linear & 0.01 \\

Phi-3-mini-128k-I   & 12,000 & 2 & 4 & 10 & 2e-4 & 4 & p\_adamw\_8bit & linear & 0.01 \\

Llama-3-8B-I + A.   & 13,000 & 2 & 4 & 5 &  2e-4 & 4 & adamw\_8bit & linear & 0.01 \\
Mistral-7B-I-v0.2 + A.   & 13,000 & 2 & 4 & 5 & 2e-4 & 4 & adamw\_8bit & linear & 0.01 \\

Llama-3-8B-I 2k   & 6,000 & 3 & 4 & 5 & 2e-4 & 4 & adamw\_8bit & linear & 0.01 \\ 
Llama-3-8B-I R 2k    & 6,000 & 3 & 4 & 5 & 2e-4 & 4 & adamw\_8bit & linear & 0.01 \\
\bottomrule
\end{tabular}
\caption{MSL means Maximum Sequence Length (Tokens), E means Epochs, GAS means Gradient Accumulation Steps, WS means Warmup Steps, LR means Learning Rate, BS means Batch Size, O means Optimizer, S means Scheduler, WD means Weight Decay. By A. Asclepius is meant. Prime (P.) means the instruction tuning runs with Asclepius.}
\label{tab:trainer}
\end{table}
The Maximum Sequence Length for the 70B models was reduced to decrease memory consumption. For Phi-3, the optimizer was changed from adamw\_8bit \citep{loshchilov2018decoupled} to paged\_adamw\_8bit\footnote{\url{https://huggingface.co/docs/bitsandbytes/en/optimizers\#paged-optimizers} Accessed: 2024-05-15} to further optimize memory usage.

\pagebreak

\section{Clinical Evaluation Criteria and Analysis}
\label{sec:app_clinic}
The submissions from the top six scoring teams underwent a comprehensive review by a team of clinicians\footnote{\url{https://stanford-aimi.github.io/discharge-me/} Accessed: 2024-05-17} at the conclusion of the competition. The generated sections were assessed based on their completeness, correctness, readability, and holistic comparison to the reference text. These criteria were evaluated on a scale ranging from 1 to 5, where 1 signifies performance that is significantly worse than the reference text, and 5 indicates performance that is significantly better than the reference text. Three independent clinicians scored 25 DI and 25 BHC texts from each team, using the same DS.

\begin{table}[h]
\centering
\begin{tabular}{lcccccccc}
\toprule
& & \multicolumn{4}{c}{BHC} & \multicolumn{3}{c}{DI} \\
\cmidrule(lr){3-6} \cmidrule(lr){7-9}
Team & Avg. & Comp & Corr & Read & Hol. & Comp & Corr & Hol.\\ 
\midrule
WisPerMed & 3.375 & 3.667 & 3.667 & 3.373 & 2.440 & 3.947 & 4.000 & 2.533 \\
HarmonAI Lab at Yale & 2.903 & 3.520 & 2.587 & 2.107 & 1.533 & 4.267 & 3.947 & 2.360 \\
aehrc & 2.785 & 2.307 & 3.053 & 1.960 & 1.093 & 3.907 & 4.547 & 2.627 \\
EPFL-MAKE & 2.720 & 3.293 & 2.827 & 2.533 & 1.653 & 3.453 & 3.413 & 1.867 \\
UF-HOBI & 2.579 & 2.480 & 3.360 & 2.707 & 1.413 & 3.013 & 3.293 & 1.787 \\
de ehren & 2.335 & 2.280 & 2.987 & 2.680 & 1.120 & 2.813 & 3.053 & 1.413 \\
\bottomrule
\end{tabular}
\caption{BHC and DI Metrics for Teams by clinicans. In this Table Avg. stands for Average, Comp stands for Comperability, Corr stands for Correctness, Read stands for Readability, and Hol stands for Holistic.}
\label{tab:metrics}
\end{table}

Notably, the ranking order of the first six teams did not change, indicating the high quality and reliability of the metrics used in the competition (see Table \ref{tab:results} and Table \ref{tab:metrics}). This consistency suggests that the chosen metrics are effective measures of the submissions' quality and performance, as they align well with clinical evaluations.

Clinician scores were normalized between 0 and 1 to make them comparable to the challenge scores. Note that readability was not compared, as clinicians did not score the readability for DI texts, nor did the challenge metrics include readability scores.

\begin{figure}[h]
\centering
  \includegraphics[scale=0.5]{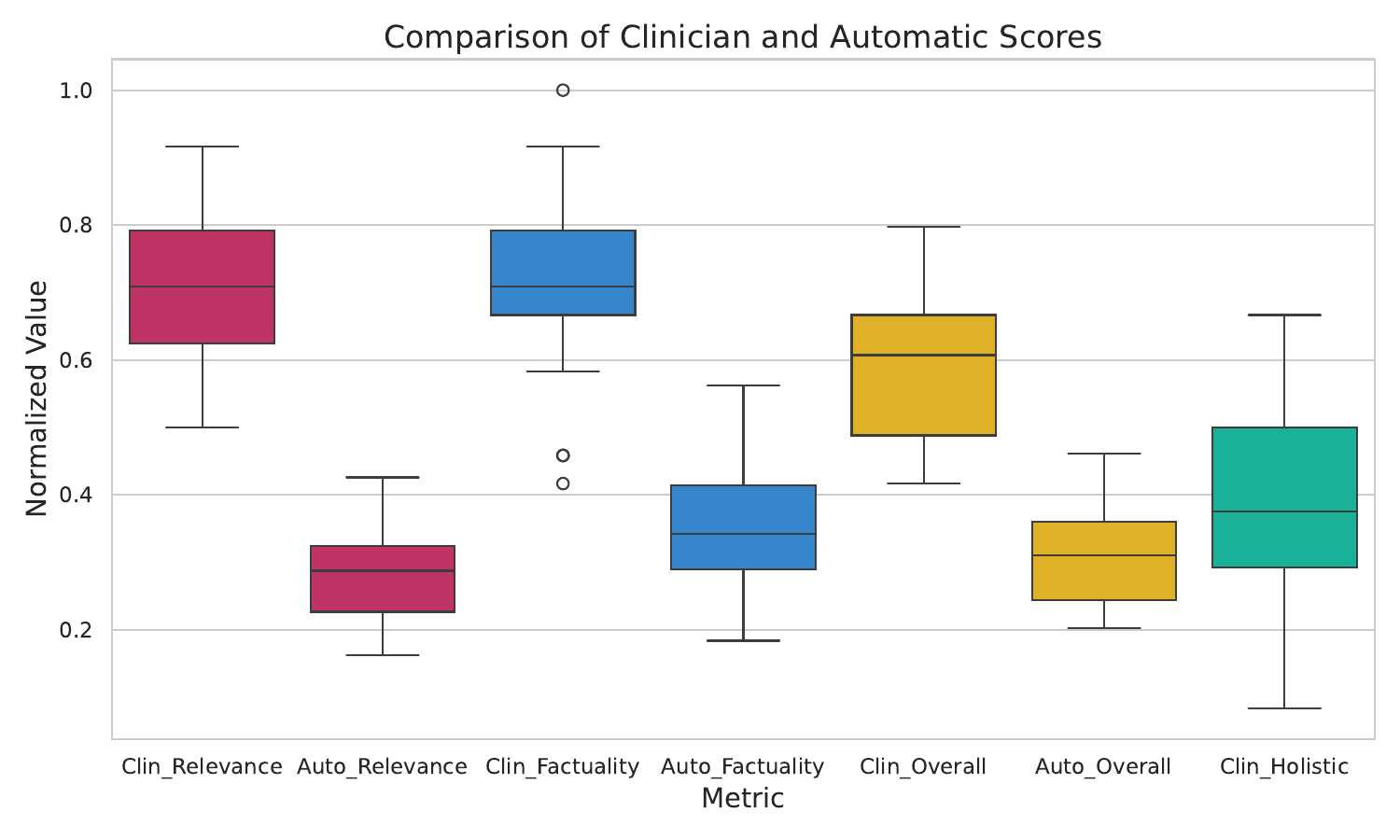}
  \caption{Boxplot of Average Clinician Scores and Average Metric Scores. Clin stands for Clinician and Auto stands for the scores caluclate with the challenge metrics. Here Auto\_Relevance includes ROUGE-1, ROUGE-2, ROUGE-L, BertScore and BLEU-4. Auto\_Factuality include AlignScore METEOR and MEDCON.}
  \label{fig:box}
\end{figure}

To conclude, while clinicians generally score higher (see Figure \ref{fig:box}) than the evaluation metrics, taking a holistic look at the entire text reveals that the average scores from the evaluation metrics align closely with those of the clinicians. This indicates that the holistic evaluation by clinicians is comparable to the automated metrics, thus validating the effectiveness of the metrics used in the competition.

\section{Environmental Impact}
\label{sec:appendix_env}

In scientific research, it is crucial to consider not only the direct results of experiments but also the broader implications and consequences of the research process. While the following environmental assessment is not directly tied to the primary results, reporting on the environmental footprint of the work is essential given the increasing global emphasis on sustainability and the environmental impact of computational practices. This perspective aligns with the findings of \citep{ulmer-etal-2022-experimental}, emphasizing the importance of understanding and reporting the environmental consequences of experimental work.\\\\
The experiments were conducted using HPC resources located in Essen and Dortmund, Germany. The region's electricity generation has a carbon efficiency of 0.381
kgCO\textsubscript{2} eq/kWh\footnote{\url{https://ourworldindatAsclepiusorg/grapher/carbon-intensity-electricity?country=~DEU} Accessed: 2024-05-14}, with approximately 41,1\% \footnote{\url{https://www.destatis.de/EN/Themes/Economic-Sectors-Enterprises/Energy/Production/Tables/gross-electricity-production.html} Accessed: 2024-05-14} of the electricity being sourced from fossil fuels. To estimate the carbon footprint of our experiments, the Machine Learning Impact calculator, as presented by \citep{DBLP:journals/corr/abs-1910-09700}, is utilized. This calculator provides a comprehensive framework to quantify the carbon emissions associated with machine learning experiments, considering both the energy consumption of computational resources and the carbon efficiency of the electricity source.

\begin{table}[h!]
\centering
\resizebox{\textwidth}{!}{%
\begin{tabular}{lcccc}
\toprule
\textbf{Final Models} & \textbf{Runtime (hours)} & \textbf{Power (Avg. Watts)} & \textbf{Energy (kWh)} & \textbf{CO\textsubscript{2} (kg)} \\
\midrule
Mistral-7B-I-v0.2 BHC + A. & 28.5 & 651.45 & 18.58 & 7.08 \\
Mistral-7B-I-v0.2 + A. Prime & 5 & 637 & 3.21 & 1.22 \\
Mistral-7B-I-v0.2 DI + A. & 27.4 & 681 & 18.71 & 7.13 \\
\midrule
\textbf{Experiment runs} & 1,920 & 783.83 & 1,511.59 & 575.91 \\
\midrule
\textbf{Overall} & 1,980.9 & 783.532 & \textbf{1,552.10} & \textbf{591.35} \\
\bottomrule
\end{tabular}%
}
\caption{Runtime, Energy Consumption and CO\textsubscript{2} Emissions for the Final models, Other Experiment Runs and Overall for All Experiments. By A. Asclepius is meant. Prime means the instruction tuning runs with Asclepius.}
\label{tab:energy_consumption}
\end{table}

The carbon footprint and electricity consumption values for our optimal models, as well as for all experimental runs conducted throughout the research process presented in table \ref{tab:energy_consumption}. The values indicate that substantial resources are expended on debugging and testing during development.

\section{Acknowledgement}
The work of Hendrik Damm, Tabea Pakull, Bahadır Eryılmaz, Helmut Becker, Ahmad Idrissi-Yaghir and Henning Schäfer was funded by a PhD grant from the DFG Research Training Group 2,535 \textit{Knowledge-and data-based personalization of medicine at the point of care (WisPerMed)}.
\newpage
\section{Licenses}
\label{sec:appendix_lic}
In Table \ref{tab:translation_metrics} the Licenses as given by the owners of the Dataset/Framework/Model are displayed. 

\begin{table}[h]
\centering
\begin{tabular}{llll}
\toprule
Dataset/Framework/Model                   & License\\ 
\midrule
Asclepius dataset\tablefootnote{\url{https://huggingface.co/datasets/starmpcc/Asclepius-Synthetic-Clinical-Notes} Accessed: 2024-05-17}   & Creative Commons Attribution Non Commercial Share Alike 4.0\\
MIMIC-IV-Note\tablefootnote{\url{https://physionet.org/content/mimic-iv-note/2.2/} Accessed: 2024-05-17} & PhysioNet Credentialed Health Data License 1.5.0\\
MIMIC-IV-ED\tablefootnote{\url{https://physionet.org/content/mimic-iv-ed/2.2/} Accessed: 2024-05-17}   & PhysioNet Credentialed Health Data License 1.5.0\\
MIMIC-SID\tablefootnote{\url{https://github.com/plandes/mimicsid} Accessed: 2024-05-17}                  & MIT License\\
unsloth\tablefootnote{\url{https://github.com/unslothai/unsloth} Accessed: 2024-05-17}                  & Apache License Version 2.0\\
Mistral-7B-I-v0.2\tablefootnote{\url{https://huggingface.co/mistralai/Mistral-7B-Instruct-v0.2} Accessed: 2024-05-17}              & Apache License Version 2.0\\
Llama-3-8B-I\tablefootnote{\url{https://huggingface.co/meta-llama/Meta-Llama-3-8B-Instruct} Accessed: 2024-05-17}                & Llama 3 Community License Agreement\\
Llama-3-70B-I\tablefootnote{\url{https://huggingface.co/meta-llama/Meta-Llama-3-70B-Instruct} Accessed: 2024-05-17}                & Llama 3 Community License Agreement\\
OpenBioLLM-70B\tablefootnote{\url{https://huggingface.co/aaditya/Llama3-OpenBioLLM-70B} Accessed: 2024-05-17}                & Llama 3 Community License Agreement\\
WizardLM-2 8x22B\tablefootnote{\url{https://huggingface.co/alpindale/WizardLM-2-8x22B} Accessed: 2024-05-17}                & MIT License\\
Phi-3-mini-128k-I\tablefootnote{\url{https://huggingface.co/microsoft/Phi-3-mini-128k-instruct} Accessed: 2024-05-17}               & Apache License Version 2.0\\ 
\bottomrule
\end{tabular}
\caption{Licenses of the dataset, Framework and Models used for this Shared Task}
\label{tab:translation_metrics}
\end{table}

\end{document}